\documentclass[10pt,twocolumn,letterpaper]{article}

\usepackage{cvpr}              
\usepackage{graphicx}
\usepackage{amsmath}
\usepackage{amssymb}
\usepackage{booktabs}
\usepackage{multirow}

\usepackage[pagebackref,breaklinks,colorlinks]{hyperref}

\usepackage[capitalize]{cleveref}
\crefname{section}{Sec.}{Secs.}
\Crefname{section}{Section}{Sections}
\Crefname{table}{Table}{Tables}
\crefname{table}{Tab.}{Tabs.}

\def\cvprPaperID{9054} 
\def\confName{CVPR}
\def\confYear{2022}

\begin{document}


\title{ClusterGNN: Cluster-based Coarse-to-Fine Graph Neural Network for Efficient Feature Matching}

\author{Yan Shi$^{1}$\footnotemark[1] \,\, Jun-Xiong Cai$^{1}$\footnotemark[1] \,\, Yoli Shavit $^{2}$\,\, Tai-Jiang Mu$^{1}$ \,\, Wensen Feng$^{3}$\footnotemark[2] \,\, Kai Zhang$^{1}$\footnotemark[2] 
\and
$^1$Tsinghua University
\and
$^2$Faculty of Engineering, Bar Ilan University
\and
$^3$Huawei Technologies
\and
\tt\small shi-y20@mails.tsinghua.edu.cn, caijunxiong000@163.com,yolisha@gmail.com
\and
\tt\small taijiang@tsinghua.edu.cn,fengwensen@huawei.com,zhangkai@sz.tsinghua.edu.cn
}

\maketitle
\renewcommand{\thefootnote}{\fnsymbol{footnote}} 
\footnotetext[1]{These authors contributed equally to this work.} 
\footnotetext[2]{Corresponding authors.}


\begin{abstract}
Graph Neural Networks (GNNs) with attention have been successfully applied for learning visual feature matching. However, current methods learn with complete graphs, resulting in a quadratic complexity in the number of features. Motivated by a prior observation that self- and cross- attention matrices converge to a sparse representation, we propose ClusterGNN, an attentional GNN architecture which operates on clusters for learning the feature matching task. Using a progressive clustering module we adaptively divide keypoints into different subgraphs to reduce redundant connectivity, and employ a coarse-to-fine paradigm for mitigating miss-classification within images. Our approach yields a 59.7\% reduction in runtime and 58.4\% reduction in memory consumption for dense detection, compared to current state-of-the-art GNN-based matching, while achieving a competitive performance on various computer vision tasks.  

\end{abstract}

\section{Introduction}
\label{sec:intro}
Finding correspondences between images is an essential task for many computer vision applications such as Simultaneous Localization and Mapping (SLAM)\cite{slam,slam2}, Structure-from-Motion (SfM)\cite{sfm,sfm2} and camera pose estimation\cite{camerapose}. Given a pair of images, correspondences can be established through point-to-point feature matching. Classical pipelines typically obtain correspondences with a nearest neighbor (NN) search of feature descriptors and reject outliers based on their match score or using a mutual NN check. Such methods focus only on the local similarity between feature descriptors while ignoring geometric information and the global receptive field. \par
\begin{figure}[t]
  \centering
  \includegraphics[width=1\linewidth]{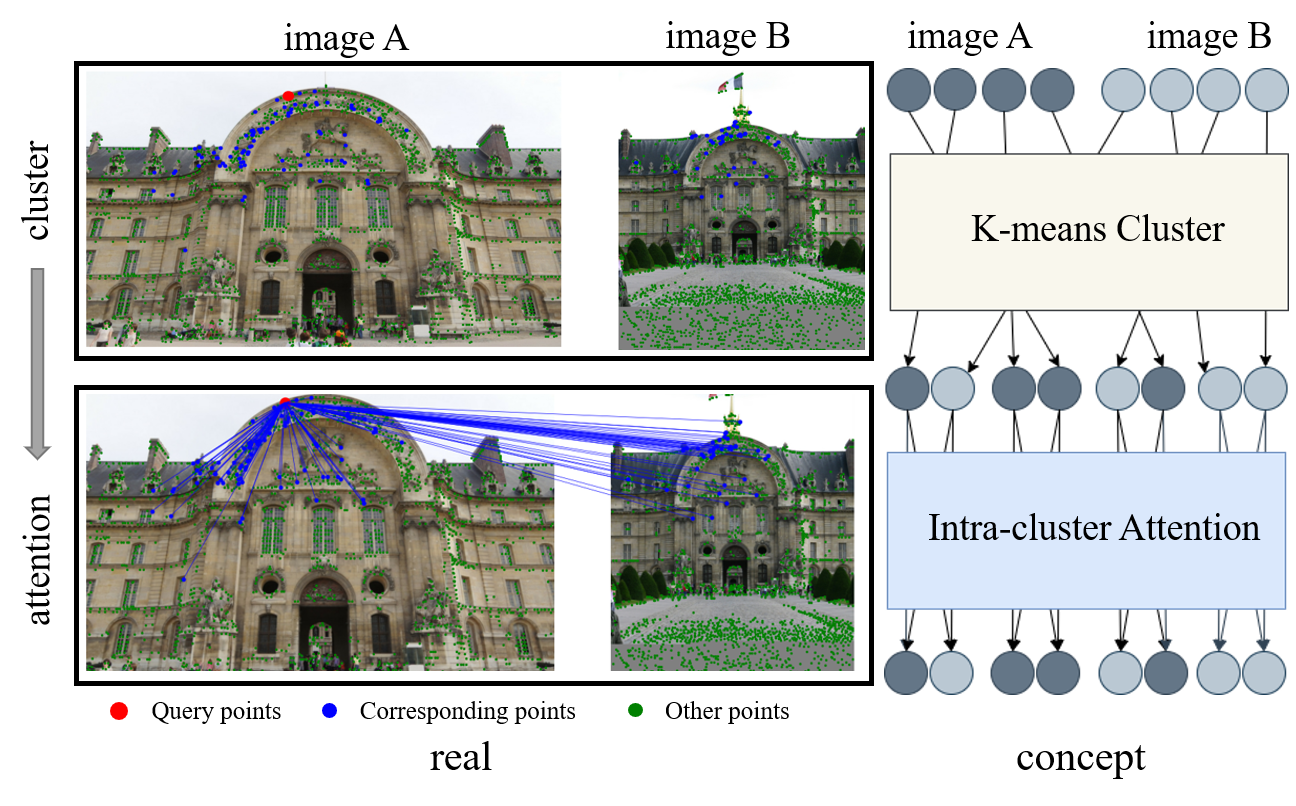}
  \caption{Sparse attention in cluster-based feature matching. Each keypoint interacts only with points within its cluster (proposed method) instead of interacting with all keypoints as in current GNN-based feature matching \cite{superglue}.}
  \label{fig:first}
\end{figure}
Recent works \cite{superglue,sgm,loftr} have proposed to learn the task of feature matching using graph neural networks (GNNs) and attention. In SuperGlue\cite{superglue}, Transformer~\cite{transformer} based GNNs are applied on the complete graph of keypoints within (intra graph) and between (inter graph) images. Each node is represented with an encoded keypoint descriptor and updated using self- and cross- multi-head attention, while alternating between the intra- and inter- complete graphs, respectively. Learning complete graphs with attention suffers from a computational and memory complexity which is quadratic in N, where N is the number of keypoints. However, keypoints typically show a strong correlation with just a small number of points (sparse adjacency matrix). Furthermore, in the context of feature matching, a large portion of keypoints are non-repeatable and irrelevant for matching. A complete graph representation is thus redundant and results in wasteful attention-based message passing. 

Efforts to reduce the quadratic complexity of attention mainly focused on self-attention. For example, reducing the attention dimension by splitting the input sequence into local windows \cite{blockattention} or by approximating attention with kernels \cite{performer}. However, these works are less suitable for feature matching, where we are required to perform self- and cross- attention on features within and between images, respectively. Inspired by the Routing Transformer \cite{routing}, we propose a coarse-to-fine cluster-based GNN to learn the feature matching task with a lower redundancy and computational complexity. We extract query and key features of keypoints across images to classify the points with strong correlation into the same cluster and establish local graphs using points from the same category. Each point interacts only with points in the same local graph resulting with a sparser attention computation (Fig. \ref{fig:first}) . Since clustering keypoints across images may lead to an erroneous grouping within images, we take a coarse-to-fine approach and first divide the points into a small number of major clusters which are gradually divided into multiple smaller clusters. We evaluate our method on multiple tasks, namely: relative pose estimation, homography estimation and visual localization. Our model achieves state-of-the-art accuracy across tasks, with a significant improvement in efficiency for dense detection (59.7\% and 58.4\% reduction in runtime and memory, respectively).\par
In summary, our contributions are as follows: 
\begin{enumerate}
    \item We present a learnable, coarse-to-fine clustering method to establish local graphs for feature matching, which reduces the spread of redundant information and makes message passing more effective.
    \item We introduce the ClusterGNN architecture, an attentional GNN for learning the feature matching task using gradually forming clusters, approximating attention on complete graphs.
    \item The proposed ClusterGNN method achieves state-of the-art results on various tasks, with a significant reduction in runtime and memory  consumption on dense detection (59.7\% and 58.4\%, respectively) compared to the current leading feature matching method (SuperGlue).
\end{enumerate}
\begin{figure*}[t]
  \centering
   \includegraphics[width=1\linewidth]{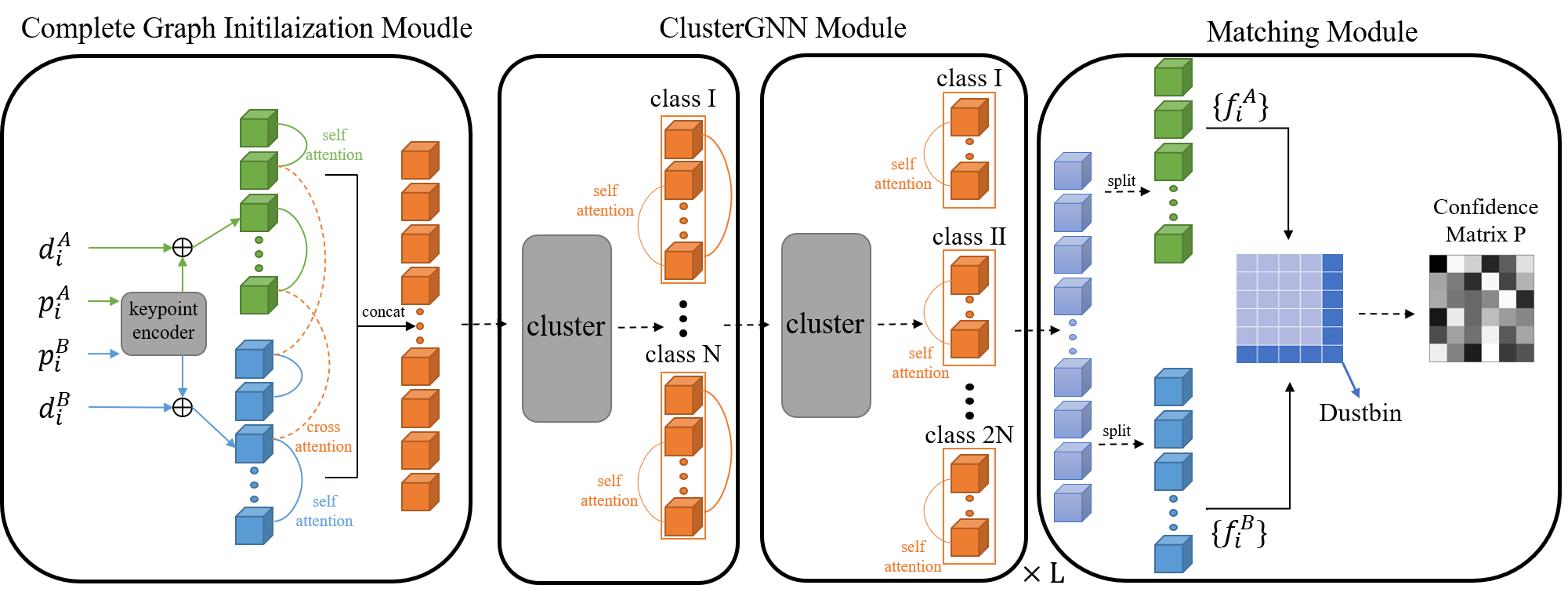}
    \caption{\textbf{The ClusterGNN architecture.} Our model is composed on three components. A Complete Graph Initialization Module (Section \ref{sec:dense}) first constructs and updates complete inter- and intra- graphs with attentional GNN layers. In order to leverage on the inherent sparsity of keypoint attention maps, a ClusterGNN Module (Section \ref{sec:clustergnn}), learns to hierarchically partition the joint complete graph into smaller sub-graphs before applying attention-base updates. Finally, a Matching Module (Section \ref{sec:matching}), establishes the matching probability matrix using dot product and the Dual Softmax operator. A learnable dustbin is further appended to account for non-matching keypoints.}
   \label{fig:main}
\end{figure*}
\section{Related Work}
\label{sec:releatedwork}
\textbf{Visual Feature matching.} Matching features between a pair of images is used in different computer vision applications, such as visual localization and relative pose estimation. In classical pipelines, keypoints are first detected and described using hand-crafted methods such as SIFT\cite{sift} and ORB\cite{orb}. The nearest neighbor of each descriptor is obtained based on Euclidean distances, followed by filtering of incorrect matches using mutual consistency check, Lowe’s ratio test\cite{sift}, matching scores and neighborhood consensus.\par 
In recent years, deep learning methods have been employed for improving different aspects of the feature matching pipeline. Learning-based feature detectors and descriptors such as SuperPoint\cite{SuperPoint}, ASLFeat\cite{aslfeat} and R2D2\cite{r2d2}, were proposed for learning more robust, discriminative and repeatable representations of keypoints. Other methods focused instead on the matching and filtering tasks \cite{superglue,outlier,auc1}. A significant improvement in matching accuracy was recently achieved by SuperGlue\cite{superglue}, an attentional GNN architecture, which establishes complete graphs over keypoints within and between images and update their representations with an attention-based message passing. Albeit, since the runtime and memory complexity of attention is quadratic in the number of nodes, operating on the complete graph does not scale well. This in turns limits usability of SuperGlue, especially when matching a large number of points. In this work, we build on the success of SuperGlue and propose a sparser alternative to reduce the propagation of redundant messages, achieving a significant decrease in runtime and memory while preserving matching performance.\par
\textbf{Efficient Attention.} The attention mechanism was popularized through the Transformer architecture\cite{transformer}, achieving state-of-the-art results across different natural language processing and computer vision tasks\cite{bert,vit,dert}. In the context of attentional GNNs, Transformers can be viewed a graph-like model operating on the complete graph of tokens, which are updated using self- and cross- attention\cite{graph,graph2}. Since Transformers (and attentional GNNs operating on complete graphs) are hampered by the quadratic complexity of attention in the sequence length (number of nodes), different methods were recently proposed to sparsify connections or linearize the attention complexity. 

In \cite{sgm}, a small set of reliable nodes are established as seeds to reduce the cost of attention. In\cite{blockattention,blockimageattention,sparseattention}, data independent or fixed sparsity patterns were proposed to bound temporal dependencies, such as local or strided attention. Drawing inspiration from CNN networks, those works suggested to apply attention within a fixed local neighborhood, which bounds the complexity, but limits the ability to establish long range interactions. In the context of feature matching, such methods may degrade performance since the distribution of keypoints is not regular, and the two matched images may have large scale and viewing angle differences. Other approaches \cite{lowranklinformer,kernels2,poolinglongformer} proposed to approximate attention by either lowering the sequence dimension through pooling or lowering the attention matrix dimension using low-rank methods. Such approximations, however, include assumptions which are less appropriate for feature matching. 

Inspired by content-based sparse attention\cite{routing,reformer}, which use data-driven methods to cluster tokens and operate within clusters, we propose a learned clustering module, which utilizes k-means to construct sub-graphs in a coarse-to-fine module, and passes messages within each local cluster graph for saving memory and computations. 

\section{Method}
\label{sec:methodology}
\subsection{Problem Definition} Given an image pair $({\mathcal{I}_a, \mathcal{I}_b})$ and extracted  keypoints $(\mathcal{K}_a, \mathcal{K}_b)$ with corresponding confidence scores $(\mathcal{C}_a, \mathcal{C}_b)$  and descriptors $(\mathcal{D}_a, \mathcal{D}_b)$, the feature matching problem is defined as pairing keypoints which match in real-world coordinates:
\begin{equation}
    \mathcal{M}_{a,b}=\{(i, j)| \| \mathcal{T}_a(\mathcal{K}_a^{(i)}) - \mathcal{T}_b(\mathcal{K}_b^{(j)}) \| \leq \epsilon \}
\end{equation}
where $\mathcal{T}_x$ represent the function which transfer pixel coordinates to world coordinates.
In practice, $\mathcal{T}_x$ is usually obtained by homography matrix estimation based on predicted matches. When faced with large
differences between camera views and long-term environmental changes, traditional methods which rely on NN search often produce wrong matches. Recent GNN-based methods apply attention in an iterative manner to remove matching-unrelated environmental noise and learn global geometric distribution information in order to achieve optimal matching and increase matching robustness.
\par
\subsection{Motivation}
\begin{figure}[h]
\centering
\subcaptionbox{ \label{fig:gt}}
    {\includegraphics[width=1\linewidth]{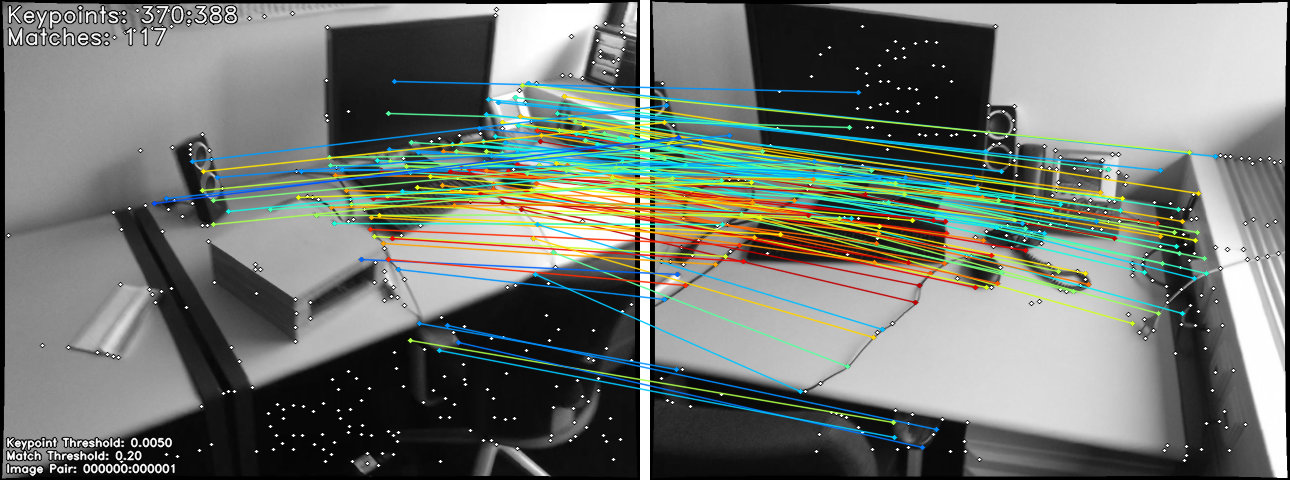}}
  \subcaptionbox{ \label{fig:matrix}}
    {\includegraphics[width=0.29\linewidth]{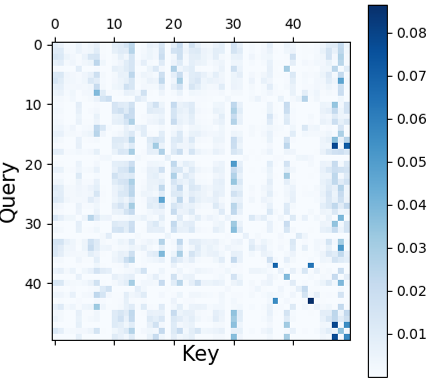}}
\subcaptionbox{\label{fig:distribution}}
    {\includegraphics[width=0.32\linewidth]{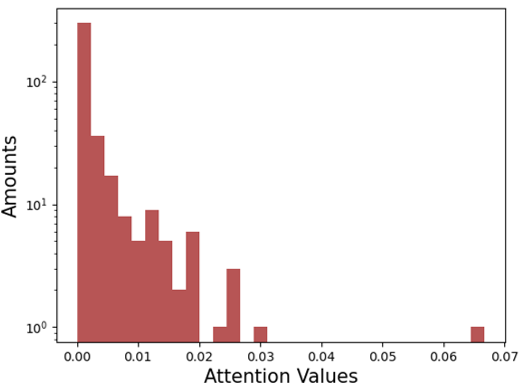}}
\subcaptionbox{ \label{fig:clustermatrix}}
    {\includegraphics[width=0.29\linewidth]{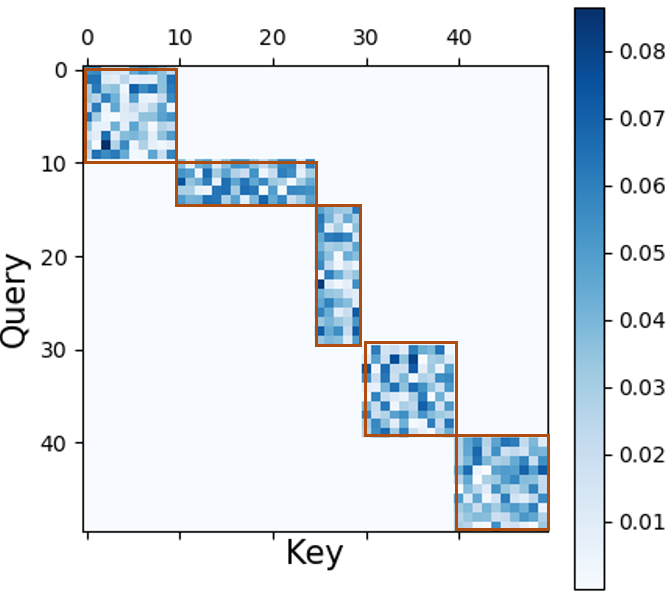}}
\caption{\textbf{(a)} Visualization of feature matching between two images. Keypoints and matches are represented by white points and colored lines, respectively where color varies between green to red according to the matching confidence (higher in red). \textbf{(b)} Cross attention matrix of 50 sampled keypoints from two images.  \textbf{(c)} The distribution of attention values in an example row in the attention matrix. \textbf{(d)} The cluster attention matrix. The attention matrix within each cluster is framed with a red box.
}
\label{fig:31}
\end{figure}

Existing works~\cite{superglue,loftr} which learn the feature matching task with attention-based GNNs, use densely connected graphs. However, as shown in Fig.\ref{fig:gt}, a large portion of keypoints are non-repeatable and irrelevant for feature matching. In addition, the respective self- and cross- attention matrices tend to converge to sparse matrices (visualized in Fig.\ref{fig:matrix}), where  most of the attention values are distributed around zero (Fig.\ref{fig:distribution}). Therefore, it is important to design a sparse structure for efficient feature matching.\par
In order to learn over sparse graphs and reduce redundancy, we extract query and key features of keypoints to classify strongly correlated points into the same cluster (Fig.\ref{fig:clustermatrix}). The points in each cluster are used to build a small sub-graph. Using attention, we can then pass information and update keypoints representation within each local sub-graph. Due to different descriptor statistics between images, direct classification may cause points from the same image to be wrongly classified into the same cluster. In order to address this problem, we propose a cluster-based coarse-to-fine paradigm and apply attentional GNN layers withing clusters for learning feature matching between two sets of feature points and their associated descriptors.\par
\subsection{Network Architecture}
Similarly to \cite{superglue}, our method first applies a \textbf{Complete Graph Initialization Module} (Section \ref{sec:dense}), which constructs complete graphs over encoded keypoints and descriptors within and between images (intra- and inter- graphs respectively), and updates them using attention. Instead of learning multiple attentional GNN layers over theses complete graphs, we design a \textbf{ClusterGNN module} (Section \ref{sec:clustergnn}), which learns to hierarchically partition the complete graph into smaller sub-graphs, and then applies the attention update mechanism within these graphs. Finally, the matching probability between keypoints is computed with a \textbf{Matching Module} (Section \ref{sec:matching}), based on the dot product of updated feature representations and the Dual-Softmax operator \cite{softmaxneighbourhood,loftr}. An overview of our proposed method is shown in Fig. \ref{fig:main}.

\subsubsection{Complete Graph Initialization Module}
\label{sec:dense}
Given input keypoints $\mathcal{K}$, confidence scores $\mathcal{C}$ and local descriptors $\mathcal{D}$, we generate a joint representation as in \cite{superglue} by adding descriptors and encoded keypoints and confidence scores. We then construct complete intra (within images) and inter (between images) graphs over the keypoints in $({\mathcal{I}_a, \mathcal{I}_b})$ and apply multi-head attention as in \cite{transformer} and \cite{superglue} for updating node representation. This construction and initialization process can be expressed as follows:
\begin{align}
    \mathcal{D}_a^0, \mathcal{D}_b^0 & = CA(SA(KE(\mathcal{I}_a)), SA(KE(\mathcal{I}_b))), \nonumber \\
     \mathcal{D}_a^i, \mathcal{D}_b^i & = CA(SA(\mathcal{D}_a^{i-1}), SA(\mathcal{D}_b^{i-1})), \nonumber \\
    KE(\mathcal{I}_x)                  & = \mathcal{D}_x + MLP(\mathcal{K}_x\oplus \mathcal{C}_x),
\end{align}
where $SA$/$CA$ stands for self/cross attention (introduced below), $KE$ is the keypoint encoding layer, and $\oplus$ denotes the concatenation operator.
For two input feature sets $(\mathcal{F}_a,\mathcal{F}_b)$, we implement attention-based GNNs to pass and aggregate messages between nodes, as follows:
\begin{align}
    \label{equ:att}
    CA(\mathcal{F}_a, \mathcal{F}_b)  & = \mathcal{F}_{a,b} + MLP(\mathcal{F}_{a,b} \oplus Att(\mathcal{F}_{a,b}, \mathcal{F}_{b,a}) ), \nonumber \\
    Att(\mathcal{F}_a, \mathcal{F}_b) & = softMax(\frac{Q_aK_b^T}{\sqrt{dim}})V_b, \nonumber                                              \\
    (Q_x,K_x,V_x)                     & = Linear_{(Q,K,V)}(\mathcal{F}_x), x\in\{a,b\},                                                   \\
    SA(\mathcal{F}_x)                 & =  CA(\mathcal{F}_x, \mathcal{F}_x),x\in\{a,b\}.
\end{align}
In practice, we use multi-head attention\cite{transformer} to improve the expressivity.
In addition, all MLP operators are followed with batch normalization and ReLU before the last layer. 

\subsubsection{Cluster GNN Module}\label{sec:clustergnn}
\textbf{Cluster-based Sparse Attention} \label{sec:sparse}
The computational bottleneck of the attentional GNNs (Equation~\ref{equ:att}) is due to the matrix multiplication operation between the query and the key matrices. However, since keypoints are likely to be correlated to only a small number of points (Fig.\ref{fig:matrix}), it is desirable to operate over local sub-graphs (where the corresponding matrices are much smaller). The Cluster-GNN module implements a cluster-based sparse attention to approximate self- and cross- attention over complete graphs, as follows:
\begin{align}
    \label{equ::satt}
    \widehat{Att}(\widetilde{\mathcal{F}})  = M \cdot  softMax(\frac{\widetilde{Q}\widetilde{K}^T}{\sqrt{dim}})\widetilde{V}, \nonumber \\
    \widetilde{\mathcal{F}}                = \mathcal{F}_a \cup \mathcal{F}_b, \nonumber \\
    \widetilde{Q} = Q_a \cup Q_b, \widetilde{K} = K_a \cup K_b, \widetilde{V} = V_a \cup V_b\nonumber \\
\end{align}
where $M=\{m_{ij}\}$ is the cluster matrix. $m_{ij}=1$ when $query_i$ (the ith \textit{query} vector of $\widetilde{\mathcal{F}}$) and $key_j$ (the jth \textit{key} vector of $\widetilde{\mathcal{F}}$) belong to the same cluster. Note that with this formulation, we no longer distinguish between features from different images (like the self/cross attention in Section~\ref{sec:dense}), and directly operate on their union.
While other works also apply the K-NN or top-k operators to determine the index matrix $M$, they do not apply well to cross attention in matching tasks.

\textbf{Learnable Hierarchical Clustering.}
\label{sec:cluster} The key to our clustering method is the definition of cluster centers and the discrepancy function.
We note that the attention map itself provides a means for measuring similarity, where the attention weights between two features are determined according to their \textit{query} and \textit{key} vectors, as follows:
\begin{align}
    \alpha_{i,j} = \frac{exp(query_i\cdot key_j)}{\sum_{k}exp(query_i\cdot key_k)}
\end{align}
 Since the \textit{query} and \textit{key} vectors are different linear projections derived from the input features (Eq.~\ref{equ:att}), $ \alpha_{i,j} \neq  \alpha_{j,i}$ is satisfied in most cases. Thus, instead of directly clustering the feature vectors, we propose to cluster the union space of the \textit{query} and \textit{key} vectors in order to get a better sparse approximation.

Given $k$ known cluster vectors $\{c_i\}$ and features $\{f_i|f_i\in Q\cup K \}$, the discrepancy function is defined as follows:
\begin{align}
    dis(c_i, f_j) = 1 - \frac{c_i\cdot f_j}{\| c_i\| \|f_j\| },
\end{align}
Using this function, we assign each feature with a cluster center index $cid$ based on its closest cluster and construct the cluster Matrix $M$ in Equation~\ref{equ::satt}.
Note that, in practice, $M$ does not need to be calculated explicitly.

During the training process, we initialize clusters centers with the k-means algorithm, and update them as follows:
\begin{equation}
    c_i = \beta c_i +(1- \beta ) (\sum_{cid_j=i} f_j)
\end{equation}
where ${\beta}$ is a weight to balance the old and new value of ${c}$, which is set to $0.99$. During test time, the trained cluster centers are directly used to cluster the input features, thus avoiding the iterative process used in traditional clustering algorithms. 

\begin{figure}[t]
  \centering
  \includegraphics[width=1\linewidth]{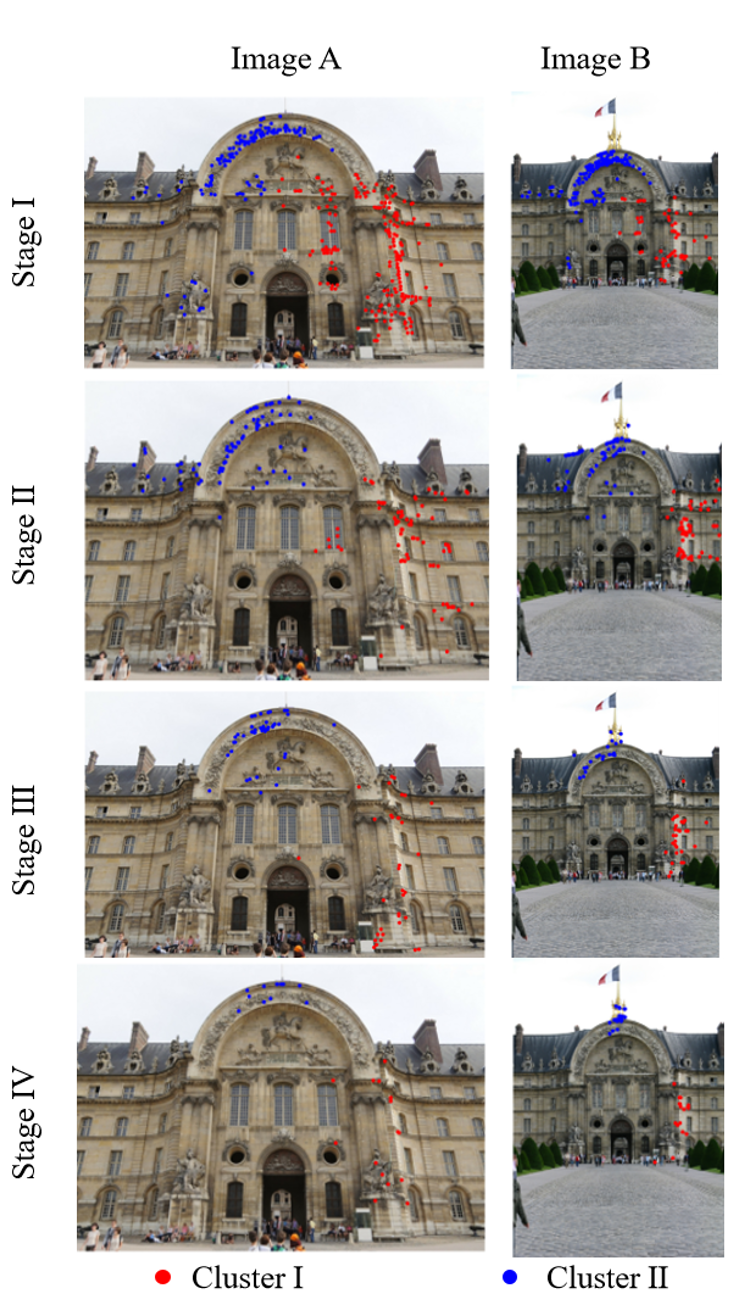}
   \caption{Visualization of hierarchical clustering. We show two clusters (colored in red and blue, respectively) at four different stages of ClusterGNN. The hierarchical clustering enables coarse-to-fine grouping }
   \label{fig:cluster}
\end{figure}

By using hierarchical clustering, we reduce the theoretical complexity of attention from $O(n^2)$ to $O(\frac{n^2}{k})$, where $n$ is the number of features and $k$ is the number of clusters. Since $k$ controls the tradeoff between the quality of approximation (better for small values of $k$) and the amount of calculations (smaller for large values of $k$), its value should be set with consideration. In theory, the best balance between clustering semantics and computational complexity can be reached when we set cluster number $k=\sqrt{n}$ \cite{routing}. However, the input features processed in \cite{routing} come from the same image and share similar statistics. For the image matching problem, the input features are keypoint descriptors extracted from two images, which may present significant style differences. Directly using the fixed $\sqrt{n}$ as the number of clusters can thus lead to over-segmentation, so that each cluster only contains features from the same image. In order to mitigate this behavior we apply clusterGNN with an increasing number of clusters, realizing a coarse-to-fine semantic clustering (Fig. ~\ref{fig:cluster}). We further carry an ablation study to guide the choice of $k$ and evaluate its effect. 

\subsubsection{Matching Module}\label{sec:matching}
We establish a matching confidence matrix $\mathcal{C}$, by applying the dot product operator between the features computed with the ClusterGNN module, $\{\mathcal{F}_a^o(i), i=1,2,...,n\}$ and $\{\mathcal{F}_b^o(i), i=1,2,...,m\}$:
\begin{equation}
   \mathcal{C} = \{\mathcal{C}_{i,j} =  \mathcal{F}_a^o(i) \cdot \mathcal{F}_b^o(j) \}
\end{equation}
\par
We further add an extra learnable dustbin channel on both row and column of the confidence matrix $\mathcal{C}$ ( $\widetilde{\mathcal{C}}$) in order to detect keypoints that are not matched. We adopt the Dual-softmax operator \cite{softmaxneighbourhood,loftr} for computing the matching probability matrix $\mathcal{P}$, by applying the log-softmax operator on both the row and column dimensions of $\widetilde{\mathcal{C}}$, as follows:
\begin{equation}
    \mathcal{P}_{i,j}=logSoftMax(\widetilde{\mathcal{C}}_{i,\cdot})_j + logSoftMax(\widetilde{\mathcal{C}}_{\cdot,j})_i.
\end{equation}
At test time, we predict the matches using the argmax operator and a mutual check mechanism.
\subsection{Loss}
The ground truth matches set $\mathcal{M}$ and non-matching keypoints set $(\overline{\mathcal{M}}_a, \overline{\mathcal{M}}_b)$ are supervised with the cross projection error (less than 3 pixels for matches and more than 5 pixels for non-matches). The matching loss ${L}_m$ is defined as:
\begin{equation}
\begin{split}
     \mathcal{L}_m= -& \left|\mathcal{M}\right| \sum_{(i,j)\in \mathcal{M} } \mathcal{P}_{ij}- \\
   &\left|\overline{\mathcal{M}}_a\right|\sum_{i\in \overline{\mathcal{M}}_a } \mathcal{P}_{i,m+1}- \left|\overline{\mathcal{M}}_b\right| \sum_{j\in \overline{\mathcal{M}}_b } \mathcal{P}_{n+1,j}
\end{split}
\end{equation}
\par
 In order to achieve a better clustering effect and enable semi-supervision, we further introduce a clustering loss: ${\mathcal{L}_c}$:
\begin{equation}
    {\mathcal{L}_c^t} = \sum_{i,j} \|c_i^t-f_j^t\|.
\end{equation}
Our total loss consists of the matching loss $\mathcal{L}_m$ and the clustering losses $\mathcal{L}_c^t$ computed across different clustering stages:
\begin{equation}
    \mathcal{L}=\mathcal{L}_m+ \gamma \sum_{t \in \left\{1,2,..,L\right\}} \mathcal{L}_c^t,
\end{equation}\par
where $\gamma$ is set to $0.1$ for balancing the two losses, $t$ indicates the clustering stage and $L$ is the total number of stages.
\subsection{Implementation Details}
\label{sec:imp}
We train our model on the MegaDepth dataset\cite{megadepth}, a large outdoor dataset, including 1M internet images from 196 different locations and sparse 3D models computed with COLMAP\cite{sfm}. It also uses multi-view stereo to generate depth maps. We select training image pairs as in \cite{d2netsfm,sfmtool1}, based on the overlap rate from the SfM co-visibility and resize images so that the larger edge is of size 1600.\par
Our model is optimized using Adam with an initial learning rate of $1 \times 10^{-4}$. We implement attention with a four heads multi-head attention throughout. Our Graph Initialization module consists of three stacked self/cross attention GNN layers. In order to reduce the GPU memory cost, we further chunk query vectors into four parts for this module. Our Cluster GNN module involves four stages with a different number of clusters, set as $\{ 16,32,64,128\}$, respectively, where each stage consists of two cluster-based GNN attention layer.
\begin{figure*}[t]
  \centering
   \includegraphics[width=1\linewidth]{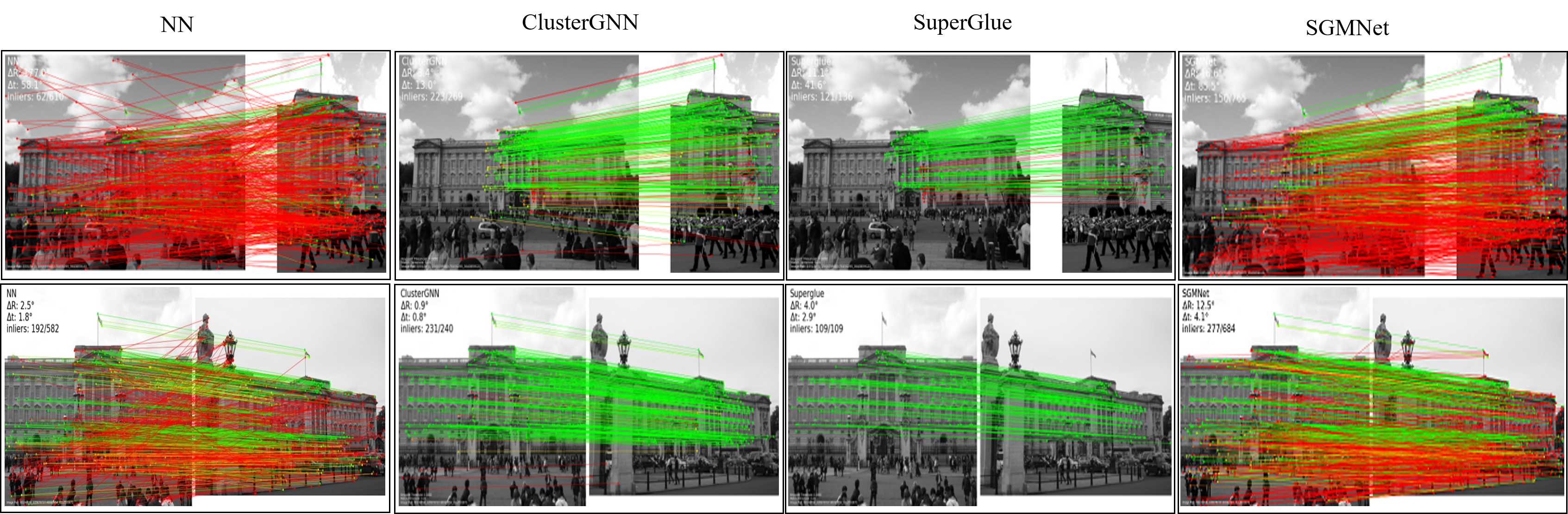}
   \caption{\textbf{Qualitative results.} The comparison of NN, Superglue, SGMNet and ClusterGNN on YFCC100M dataset with ASLFeat~\cite{aslfeat}. The number of keypoints is 2048. The red color indicates outliers and green color indicates inliers. ClusterGNN obtains more correct matches.}
   \label{fig:correspondence}
\end{figure*}
\section{Experiments}
\label{sec:Experiments}
Feature matching is a challenging task due to different factors such as occlusions, illumination and weather changes. We evaluate ClusterGNN on three different tasks, which heavily rely of feature matching, namely: pose estimation, homography estimation and visual localization. 

We compare our method with NN-search, SGMNet\cite{sgm} and SuperGlue \cite{superglue}, using both hand crafted features (SIFT \cite{sift}) and learning-based features (ASLFeat\cite{aslfeat} and SuperPoint\cite{SuperPoint}). For SGMNet, we retrain it ASLFeat using the official training code. For SuperGlue, since the official training code is not available and since its public model (denoted as Superglue$^*$) was trained on MegaDepth\cite{megadepth}, Oxford and Paris datasets \cite{oxford}), we retrain it with different features following the training method described in the SuperGlue paper. We report both our own implementation and the official results reported in the SuperGlue paper. We also provide analysis of computation and memory efficiency. All the reported experiments were run on a Tesla P-100 GPU with 16GB memory.

\subsection{Pose Estimation}
\label{sec:outdoorpose}
 We use the YFCC100M\cite{yfcc} dataset to evaluate the performance of ClusterGNN on the pose estimation task. This dataset provides a sparse reconstruction from SfM \cite{sfm,sfmtool1,sfmtool2} and ground truth poses. Following \cite{auc1,auc2,auc3,superglue}, we report the AUC of pose errors at different thresholds ($5^{\circ},10^{\circ},20^{\circ}$), where the pose errors are computed by the maximum angular differences between estimated results and ground truth in rotation and translation. For pose estimation, we use RANSAC as a post-processing tool to compute the essential matrix from predicted matches. \par
 As shown in Table \ref{tab:outdoorpose}, ClusterGNN outperforms other methods with ASLFeat. When using SuperPoint, our method outperforms our re-implemented SuperGlue and SGMNet, with a slight degradation compared to the official version. 
 When using SIFT, our method outperforms SuperGlue, with a slight degradation compared to SGMNet.
These results demonstrate the approximation and denoising abilities of ClusterGNN. Qualitative inspection (Fig. \ref{fig:correspondence}), further shows that ClusterGNN detects a larger number of true matches compared to other methods, resulting in an improved pose estimation. 
\begin{table}
\centering
\setlength{\tabcolsep}{8pt}
\resizebox{8cm}{!}{
\begin{tabular}{ccccc}
\toprule
\multicolumn{1}{c}{ {Local} }& \multicolumn{1}{c}{ \multirow{2}{*}{{Matcher}} } & \multicolumn{3}{c}{{Pose estimation}} \\
\cmidrule{3-5}
Features&&{$5^\circ$}&{$10^\circ$}&{$20^\circ$}\\
\midrule
\multirow{3}{*}{SIFT\cite{sift}}&NN&15.19&24.72&35.30\\
&SGMNet&\textbf{35.63}&\textbf{55.40}&\textbf{71.95}\\
&SuperGlue&30.12&47.25&63.05\\
&ClusterGNN&32.82&50.25&65.89\\
\midrule
\multirow{4}{*}{SuperPoint\cite{SuperPoint}}&NN&14.12&28.86&44.90\\
&SGMNet&32.44&52.80&69.90 \\
&SuperGlue$^*$&\textbf{39.02}&\textbf{59.51}&\textbf{75.72}\\
&SuperGlue&34.93&55.31&72.34\\
&ClusterGNN&35.31&56.13 &73.56 \\
\midrule
\multirow{3}{*}{ASLFeat\cite{aslfeat}}&NN&14.40&27.80&   43.20 \\
&SGMNet&32.22&52.53&70.16\\
&SuperGlue&37.50 &58.30&75.07 \\
&\textbf{ClusterGNN}&\textbf{42.62}&\textbf{61.22}&\textbf{76.75} \\
\bottomrule
\end{tabular}}
\caption{\textbf{Pose estimation using the YFCC100M dataset.} We compare different matching methods (NN, SuperGlue, SGMNet and ClusterGNN) across different keypoint extraction methods (SIFT, ASLFeat and SuperPoint). We report the AUC of pose errors at different thresholds (Section~\ref{sec:outdoorpose}). {Superglue$^*$} indicates the officially released model. The best performance is highlighted in bold.}
\label{tab:outdoorpose}
\end{table}

\subsection{Homography Estimation}
We evaluate our method on the homography estimation task using the HPathces dataset \cite{hpatches}. HPatches consist of 52 sequences that exhibit large illumination changes and 56 sequences under significant viewpoint changes. We follow the setup proposed in Patch2Pix\cite{patch2pix}, and report the percentage of correctly estimated homographies whose average corner error distance is below 1/3/5 pixels. For all compared methods, we apply the OpenCV RANSAC toolbox to estimate homography matrix. The local keypoints of SuperPoint\cite{SuperPoint} and ASLFeat\cite{aslfeat} are both tested in detail. To make a fair comparison, we choose 4k keypoints for all methods and use the same hyper-parameters.
\par
As shown in Table \ref{tab:hpatch}, ClusterGNN achieves competitive performance with a slight improvement compared to SuperGlue and SGMNet.\par
\begin{table}
\centering
\setlength{\tabcolsep}{8pt}
\resizebox{8.2cm}{!}{
\begin{tabular}{ccccc}
\toprule
\multicolumn{1}{c}{ {Local} }& \multicolumn{1}{c}{ \multirow{2}{*}{{Matcher}} } & overall & Illumination & viewpoint \\
\cmidrule{3-5}
Features&&\multicolumn{3}{c}{Accuracy($\% ,\epsilon < 1/3/5 px$) } \\
\midrule
\multirow{4}{*}{SuperPoint\cite{SuperPoint}}&NN & {0.46 / 0.78 / 0.85}&{0.57 / 0.92 / 0.97}&{0.35 / 0.65 / 0.74}\\
&SGMNet&{\textbf{0.52 / 0.85 / 0.91}} &{ 0.59 / \textbf{ 0.94 / 0.98 }}&{\textbf{0.46 / 0.74 / 0.84}}\\
&SuperGlue & {0.51 / 0.83 / 0.89}& {\textbf{0.61} / 0.93 / \textbf{0.98}} &{\textbf{0.45} / 0.73 / \textbf{0.83}}\\
&\textbf{ClusterGNN}& {\textbf{0.52} / 0.84 / 0.90}&{\textbf{0.61} / 0.93 / \textbf{0.98}}&{0.44 / \textbf{0.74} / 0.81}\\
\midrule
\multirow{4}{*}{ASLFeat\cite{aslfeat}}&NN &{0.48 / 0.81 / 0.88}&{0.57 / 0.92 / 0.97}&{0.34 / 0.68 / 0.78}  \\
&SGMNet &{0.49 / 0.82 / \textbf{0.89}}&{0.57 / 0.93 / 0.98}&{0.41 / 0.72/ \textbf{0.83}} \\
&SuperGlue &{0.49 / \textbf{0.83 / 0.89}}&{0.57 / 0.92 / \textbf{0.98}}&{0.41 / \textbf{0.72} / 0.81} \\
&\textbf{ClusterGNN} &{\textbf{0.51 / 0.83 / 0.89}}&{\textbf{0.61 / 0.95 / 0.98}}&{\textbf{0.42 / 0.72} / 0.82} \\
\bottomrule
\end{tabular}}
\caption{\textbf{Homography Estimation on Hpatches.} We report the percentage of correctly estimated homographies under different corner error distances.}
\label{tab:hpatch}
\end{table}

\subsection{Visual localization}
Visual localization is one of the most important applications of feature matching and its performance heavily relies on the matching quality. Given a query image, visual localization aims to estimate its 6-DOF position, based on a 3D reconstructed model. We integrate our method into the official HLoc\cite{hloc} pipeline for visual localization and evaluate it on the Long-Term Visual Localization Benchmark\cite{benchmark}. This benchmark assesses performance under different conditions, such as texture-less scenes of indoor environments and day-and-night changes, thus requiring highly robust matching. Specifically, we use the Aachen Day-Night dataset\cite{aachen1,aachen2} for evaluating outdoor localization and the InLoc dataset\cite{inloc} for evaluating indoor localization. The Aachen Day-Night dataset provides 4328 images of the Aachen city and 922 query images including 824 daytime images and 98 nighttime images taken by mobile phone cameras. The InLoc dataset\cite{inloc} offers 9972 reference images and 329 query images which contain significant occlusions and variation in viewpoint and illumination. \par

We use the official HLoc\cite{hloc} pipeline for visual localization tasks. Consistent with the official benchmark, we report the pose estimation accuracy under different thresholds. Tables \ref{tab:indoor} and \ref{tab:outdoor} report the results for indoor and outdoor localization, respectively. ClusterGNN achieves competitive performance compared to Superglue on both indoor and outdoor localization tasks, across different features. 
\begin{table}[h]
\centering
\setlength{\tabcolsep}{8pt}
\resizebox{8cm}{!}{
\begin{tabular}{cccc}
\toprule
\multicolumn{1}{c}{ {Local} }& \multicolumn{1}{c}{ \multirow{2}{*}{{Matcher}} } & DUC1 &DUC2\\
\cmidrule{3-4}
Features&&\multicolumn{2}{c}{(0.25m,10$^\circ$) / (0.5m,10$^\circ$) / (1.0m,10$^\circ$)} \\
\midrule
\multirow{4}{*}{SuperPoint\cite{SuperPoint}}&NN & {40.4 / 58.1 / 69.7}&{42.0 / 58.8 / 69.5} \\
&SGMNet &{41.9 / 64.1 / 73.7}&{39.7 / 62.6 / 67.2}\\
&Superglue & {\textbf{49.0} / 68.7 / \textbf{80.8}}&{\textbf{53.4} / \textbf{77.1} / 82.4}\\
&\textbf{ClusterGNN}& 47.5 / \textbf{69.7} / 79.8&\textbf{53.4} /  \textbf{77.1} / \textbf{84.7} \\
\midrule
\multirow{4}{*}{ASLFeat\cite{aslfeat}}&NN & 39.9 / 59.1 / 71.7&43.5 / 58.8 / 64.9 \\
&SGMNet &{43.9 / 62.1 / 68.2}&{45.0 / 63.4 / 73.3}\\
&Superglue&51.5 / 66.7 / 75.8 &53.4 / \textbf{76.3 / 84.0} \\
&\textbf{ClusterGNN} &\textbf{52.5 / 68.7 / 76.8} &\textbf{55.0} / 76.0 / 82.4 \\
\bottomrule
\end{tabular}}
\caption{\textbf{Indoor Localization Results (InLoc Dataset).} We report the percentage of correctly localized queries under different thresholds. The evaluation metrics are defined according to the leaderboard of Long-Term Visual Localization Benchmark\cite{benchmark}.}
\label{tab:indoor}
\end{table}

\begin{table}[h]
\centering
\setlength{\tabcolsep}{8pt}
\resizebox{8cm}{!}{
\begin{tabular}{cccc}
\toprule
\multicolumn{1}{c}{ {Local} }& \multicolumn{1}{c}{ \multirow{2}{*}{{Matcher}} } & Day &Night\\
\cmidrule{3-4}
Features&&\multicolumn{2}{c}{(0.25m,2$^\circ$) / (0.5m,5$^\circ$) / (1.0m,10$^\circ$)} \\
\midrule
\multirow{4}{*}{SuperPoint\cite{SuperPoint}}&NN & {85.4 / 93.3 / 97.2}&{75.5 / 86.7 / 92.9} \\
&SGMNet &{86.8 / 94.2 / 97.7}&{83.7 / 91.8 / 99.0} \\
&Superglue & {\textbf{89.6} / 95.4 / \textbf{98.8}}&{86.7\textbf{ / 93.9 / 100.0}}\\
&\textbf{ClusterGNN}&{89.4 / \textbf{95.5} / 98.5}&{81.6\textbf{ / 93.9 / 100.0}} \\
\midrule
\multirow{4}{*}{ASLFeat\cite{aslfeat}}&NN & 82.3 / 89.2 / 92.7 & 67.3 / 79.6 / 85.7\\
&SGMNet &{86.8 / 93.4 / 97.1}&{\textbf{86.7}/ 94.9 / 98.0}\\
&Superglue &{87.9 / 95.4 / 98.3}&{81.6 / 91.8 / 99.0} \\
&\textbf{ClusterGNN} &{\textbf{88.6 / 95.5 / 98.4}} &{85.7 \textbf{/ 93.9 / 99.0}} \\
\bottomrule
\end{tabular}}
\caption{\textbf{Indoor Localization Results (Aachen Day-Night Benchmark(v1.0)).} The evaluation metrics are refer to the leaderboard of Long-Term Visual Localization Benchmark\cite{benchmark}.}
\label{tab:outdoor}
\end{table}

\subsection{Efficiency}
Improving the processing efficiency and GPU memory requirements, while maintaining or improving matching performance is the main motivation for developing ClusterGNN.  In this section, we compare the runtime and memory of our method with SGMNet and SuperGlue.\par
\begin{figure}
\centering
\subcaptionbox{The comparison of time \label{fig:time}}
    {\includegraphics[width=0.49\linewidth]{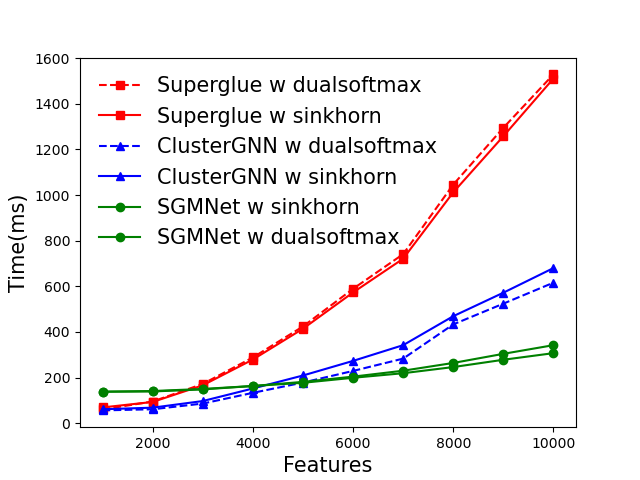}}
  \subcaptionbox{The comparison of memory\label{fig:memory}}
    {\includegraphics[width=0.49\linewidth]{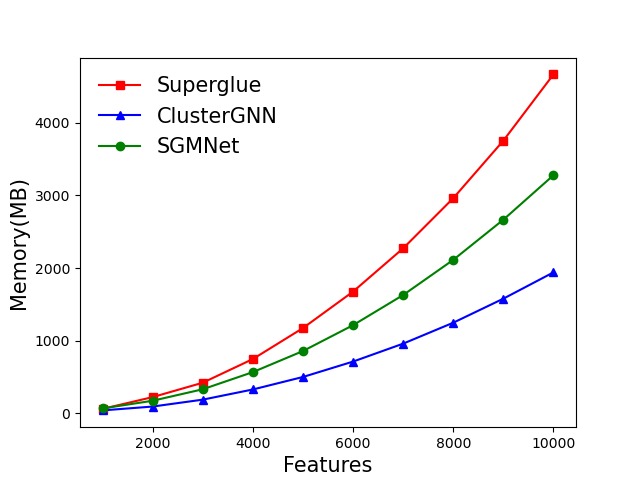}}
\caption{\textbf{Efficiency Comparison.} We report the time and memory consumption with an increasing number of input keypoints.
}
\label{fig:consumption}
\end{figure}

 
Fig. \ref{fig:time} and \ref{fig:memory}, report the time  and memory consumption for different numbers of detected keypoints on a Telsa P-100 GPU with 16GB memory. Specifically, we test both run time with Sinkhorn iterations and dual softmax. It should be noted that memory consumption is consistent in both Sinhorn and dual softmax.
As shown in Fig. \ref{fig:time}, for 10k keypoints(dense detection), ClusterGNN reduces the runtime by 59.7\%, compared to Superglue when using dual softmax. Although SGMNet is more time efficient, ClusterGNN can achieve better memory efficiency and performances at the same time. As shown in Fig. \ref{fig:memory},for dense detection, our method the proposed reduces 58.4\% of the memory required by Superglue.
\subsection{Ablation Study}
\label{sec:ablation}
 We conduct ablation studies on the YFCC100M dataset using ASLFeat~\cite{aslfeat} for detecting keypoints and the same experimental setting as in Section \ref{sec:outdoorpose}. Our ablation focuses on the effect of fixing or varying the number of clusters and on the operator used for generating the matching probability matrix. \par
\textbf{Fixed vs. Varying Number of Clusters.} Our ClusterGNN strategy gradually decreases the size of clusters while increasing their number due to the tradeoff between over segmentation and efficiency. In this experiment we evaluate the effect of fixing the number of clusters versus our choice to gradually increase it. Table \ref{tab:fixed} shows the results of fixing the number of cluster to \{16, 32, 64, 128\} versus the results of our varying strategy. ClusterGNN with a fixed number of cluster suffers from significant drop in performance. \par
\textbf{Sinkhorn vs Dual Softmax.}
Superglue\cite{superglue} uses a differential iterative optimal transport,namely Sinkhorn\cite{sinkhorn1967concerning,cuturi2013sinkhorn}, to solve the matching confidence. In our work, we adopt non-iterative dual-softmax ~\cite{softmaxneighbourhood} operator for efficiency. As shown in Table \ref{tab:sinkvsds}, dual-softmax achieves competitive performance compared with Sinkhorn.
\begin{table}
\centering
\setlength{\tabcolsep}{8pt}
\resizebox{6cm}{!}{
\begin{tabular}{cccc}
\toprule
 \multicolumn{1}{c}{ \multirow{2}*{Methods} } & \multicolumn{3}{c}{Pose estimation} \\
\cmidrule{2-4}
&{$5^\circ$}&{$10^\circ$}&{$20^\circ$}\\
\midrule
Fixed Cluster:16 &31.26&48.55&64.50 \\
Fixed Cluster:32 &37.05&56.54&72.90\\
Fixed Cluster:64 &37.44&56.58&72.68\\
Fixed Cluster:128 &36.77&55.90&72.08 \\
\textbf{ClusterGNN}& \textbf{42.62}&\textbf{61.22}&\textbf{76.75} \\
\bottomrule
\end{tabular}}
\caption{\textbf{The effect of fixing or varying the number of clusters).} We report pose accuracy under different thresholds based on YFCC100M using 2K keypoints.}
\label{tab:fixed}
\end{table}

\begin{table}
\centering
\setlength{\tabcolsep}{8pt}
\resizebox{8cm}{!}{
\begin{tabular}{cccccc}
\toprule
 \multicolumn{1}{c}{ \multirow{2}*{Methods} } & \multicolumn{3}{c}{Pose estimation} & Time & Memory \\
 \cmidrule{2-4}
&{$5^\circ$}&{$10^\circ$}&{$20^\circ$}&(ms)&(MB)\\
\midrule
ClusterGNN $w$ sinkhorn&38.83&58.87&75.11& 68& \textbf{94}  \\
ClusterGNN $w$ dual softmax& \textbf{42.62}&\textbf{61.22}&\textbf{76.75} & \textbf{61} & \textbf{94} \\
\bottomrule
\end{tabular}
}
\caption{\textbf{The effect of different operators for computing matching probability.} We report pose accuracy, processing time and GPU memory for the YFCC100M dataset using 2K keypoints.}
\label{tab:sinkvsds}
\end{table}
\section{Conclusion}
In this work, we have addressed the quadratic complexity of GNN methods for feature matching. The proposed method, named ClusterGNN, leverages on the inherent sparsity of self- and cross- attention between keypoints in the complete graph and dynamically constructs local sub-graphs through a learned coarse-to-fine clustering. Extensive evaluation on several computer vision tasks demonstrates the effectiveness of our approach, achieving a competitive performance while reducing runtime and memory by 59.7\% and 58.4\%, respectively.


\newpage
{\small
\bibliographystyle{ieee_fullname}
\bibliography{ref}

\begin{thebibliography}{10}\itemsep=-1pt

\bibitem{hpatches}
Vassileios Balntas, Karel Lenc, Andrea Vedaldi, and Krystian Mikolajczyk.
\newblock Hpatches: A benchmark and evaluation of handcrafted and learned local
  descriptors.
\newblock In {\em CVPR}, pages 5173--5182, 2017.

\bibitem{poolinglongformer}
Iz Beltagy, Matthew~E Peters, and Arman Cohan.
\newblock Longformer: The long-document transformer.
\newblock {\em CoRR}, abs/2004.05150, 2020.

\bibitem{auc3}
Eric Brachmann and Carsten Rother.
\newblock Neural-guided ransac: Learning where to sample model hypotheses.
\newblock In {\em CVPR}, pages 4322--4331, 2019.

\bibitem{dert}
Nicolas Carion, Francisco Massa, Gabriel Synnaeve, Nicolas Usunier, Alexander
  Kirillov, and Sergey Zagoruyko.
\newblock End-to-end object detection with transformers.
\newblock In {\em ECCV}, pages 213--229, 2020.

\bibitem{sgm}
Hongkai Chen, Zixin Luo, Jiahui Zhang, Lei Zhou, Xuyang Bai, Zeyu Hu, Chiew-Lan
  Tai, and Long Quan.
\newblock Learning to match features with seeded graph matching network.
\newblock In {\em CVPR}, pages 6301--6310, 2021.

\bibitem{sparseattention}
Rewon Child, Scott Gray, Alec Radford, and Ilya Sutskever.
\newblock Generating long sequences with sparse transformers.
\newblock {\em CoRR}, abs/1904.10509, 2019.

\bibitem{kernels2}
Krzysztof Choromanski, Valerii Likhosherstov, David Dohan, Xingyou Song,
  Andreea Gane, Tamas Sarlos, Peter Hawkins, Jared Davis, David Belanger, Lucy
  Colwell, et~al.
\newblock Masked language modeling for proteins via linearly scalable
  long-context transformers.
\newblock {\em CoRR}, abs/2006.03555, 2020.

\bibitem{performer}
Krzysztof Choromanski, Valerii Likhosherstov, David Dohan, Xingyou Song,
  Andreea Gane, Tamas Sarlos, Peter Hawkins, Jared Davis, Afroz Mohiuddin,
  Lukasz Kaiser, et~al.
\newblock Rethinking attention with performers.
\newblock {\em arXiv preprint arXiv:2009.14794}, 2020.

\bibitem{cuturi2013sinkhorn}
Marco Cuturi.
\newblock Sinkhorn distances: Lightspeed computation of optimal transport.
\newblock {\em Advances in neural information processing systems},
  26:2292--2300, 2013.

\bibitem{SuperPoint}
Daniel DeTone, Tomasz Malisiewicz, and Andrew Rabinovich.
\newblock Superpoint: Self-supervised interest point detection and description.
\newblock In {\em CVPRW}, pages 224--236, 2018.

\bibitem{bert}
Jacob Devlin, Ming{-}Wei Chang, Kenton Lee, and Kristina Toutanova.
\newblock {BERT:} pre-training of deep bidirectional transformers for language
  understanding.
\newblock In {\em NAACL-HLT}, pages 4171--4186, 2019.

\bibitem{vit}
Alexey Dosovitskiy, Lucas Beyer, Alexander Kolesnikov, Dirk Weissenborn,
  Xiaohua Zhai, Thomas Unterthiner, Mostafa Dehghani, Matthias Minderer, Georg
  Heigold, Sylvain Gelly, Jakob Uszkoreit, and Neil Houlsby.
\newblock An image is worth 16x16 words: Transformers for image recognition at
  scale.
\newblock In {\em ICLR}, 2021.

\bibitem{d2netsfm}
Mihai Dusmanu, Ignacio Rocco, Tomas Pajdla, Marc Pollefeys, Josef Sivic,
  Akihiko Torii, and Torsten Sattler.
\newblock D2-net: A trainable cnn for joint detection and description of local
  features.
\newblock In {\em CVPR}, pages 8092--8101, 2019.

\bibitem{slam2}
Jakob Engel, Vladlen Koltun, and Daniel Cremers.
\newblock Direct sparse odometry.
\newblock {\em TPAMI}, 40(3):611--625, 2017.

\bibitem{graph}
Matthias Fey, Jan~E Lenssen, Christopher Morris, Jonathan Masci, and Nils~M
  Kriege.
\newblock Deep graph matching consensus.
\newblock In {\em ICLR}, 2020.

\bibitem{reformer}
Nikita Kitaev, {\L}ukasz Kaiser, and Anselm Levskaya.
\newblock Reformer: The efficient transformer.
\newblock In {\em ICLR}, 2020.

\bibitem{megadepth}
Zhengqi Li and Noah Snavely.
\newblock Megadepth: Learning single-view depth prediction from internet
  photos.
\newblock In {\em CVPR}, pages 2041--2050, 2018.

\bibitem{sift}
David~G Lowe.
\newblock Distinctive image features from scale-invariant keypoints.
\newblock {\em IJCV}, 60(2):91--110, 2004.

\bibitem{aslfeat}
Zixin Luo, Lei Zhou, Xuyang Bai, Hongkai Chen, Jiahui Zhang, Yao Yao, Shiwei
  Li, Tian Fang, and Long Quan.
\newblock Aslfeat: Learning local features of accurate shape and localization.
\newblock In {\em CVPR}, pages 6589--6598, 2020.

\bibitem{slam}
Raúl Mur-Artal, J.~M.~M. Montiel, and Juan~D. Tardós.
\newblock Orb-slam: A versatile and accurate monocular slam system.
\newblock {\em IEEE Transactions on Robotics}, 31(5):1147--1163, 2015.

\bibitem{sfmtool1}
Yuki Ono, Eduard Trulls, Pascal Fua, and Kwang~Moo Yi.
\newblock Lf-net: Learning local features from images.
\newblock In {\em NIPS}, pages 6237--6247, 2018.

\bibitem{blockimageattention}
Niki Parmar, Ashish Vaswani, Jakob Uszkoreit, Lukasz Kaiser, Noam Shazeer,
  Alexander Ku, and Dustin Tran.
\newblock Image transformer.
\newblock In {\em ICML}, pages 4055--4064. PMLR, 2018.

\bibitem{blockattention}
Jiezhong Qiu, Hao Ma, Omer Levy, Scott Wen-tau Yih, Sinong Wang, and Jie Tang.
\newblock Blockwise self-attention for long document understanding.
\newblock In {\em EMNLP}, pages 2555--2565, 2019.

\bibitem{oxford}
Filip Radenovi{\'c}, Ahmet Iscen, Giorgos Tolias, Yannis Avrithis, and
  Ond{\v{r}}ej Chum.
\newblock Revisiting oxford and paris: Large-scale image retrieval
  benchmarking.
\newblock In {\em CVPR}, pages 5706--5715, 2018.

\bibitem{r2d2}
Jerome Revaud, Philippe Weinzaepfel, C{\'e}sar De~Souza, Noe Pion, Gabriela
  Csurka, Yohann Cabon, and Martin Humenberger.
\newblock R2d2: repeatable and reliable detector and descriptor.
\newblock In {\em NIPS}, pages 12405--12415, 2019.

\bibitem{softmaxneighbourhood}
Ignacio Rocco, Mircea Cimpoi, Relja Arandjelovi{\'c}, Akihiko Torii, Tomas
  Pajdla, and Josef Sivic.
\newblock Neighbourhood consensus networks.
\newblock In {\em NIPS}, pages 1658--1669, 2018.

\bibitem{routing}
Aurko Roy, Mohammad Saffar, Ashish Vaswani, and David Grangier.
\newblock Efficient content-based sparse attention with routing transformers.
\newblock {\em Transactions of the Association for Computational Linguistics},
  9:53--68, 2021.

\bibitem{orb}
Ethan Rublee, Vincent Rabaud, Kurt Konolige, and Gary Bradski.
\newblock Orb: An efficient alternative to sift or surf.
\newblock In {\em ICCV}, pages 2564--2571. Ieee, 2011.

\bibitem{hloc}
Paul-Edouard Sarlin, Cesar Cadena, Roland Siegwart, and Marcin Dymczyk.
\newblock From coarse to fine: Robust hierarchical localization at large scale.
\newblock In {\em CVPR}, pages 12716--12725, 2019.

\bibitem{superglue}
Paul-Edouard Sarlin, Daniel DeTone, Tomasz Malisiewicz, and Andrew Rabinovich.
\newblock Superglue: Learning feature matching with graph neural networks.
\newblock In {\em CVPR}, pages 4938--4947, 2020.

\bibitem{aachen1}
Torsten Sattler, Tobias Weyand, Bastian Leibe, and Leif Kobbelt.
\newblock Image retrieval for image-based localization revisited.
\newblock In {\em BMVC}, volume~1, page~4, 2012.

\bibitem{sfm}
Johannes~L Schonberger and Jan-Michael Frahm.
\newblock Structure-from-motion revisited.
\newblock In {\em CVPR}, pages 4104--4113, 2016.

\bibitem{sfmtool2}
Johannes~L Sch{\"o}nberger, Enliang Zheng, Jan-Michael Frahm, and Marc
  Pollefeys.
\newblock Pixelwise view selection for unstructured multi-view stereo.
\newblock In {\em ECCV}, pages 501--518. Springer, 2016.

\bibitem{sinkhorn1967concerning}
Richard Sinkhorn and Paul Knopp.
\newblock Concerning nonnegative matrices and doubly stochastic matrices.
\newblock {\em Pacific Journal of Mathematics}, 21(2):343--348, 1967.

\bibitem{loftr}
Jiaming Sun, Zehong Shen, Yuang Wang, Hujun Bao, and Xiaowei Zhou.
\newblock Loftr: Detector-free local feature matching with transformers.
\newblock In {\em CVPR}, pages 8922--8931, 2021.

\bibitem{outlier}
Weiwei Sun, Wei Jiang, Eduard Trulls, Andrea Tagliasacchi, and Kwang~Moo Yi.
\newblock Acne: Attentive context normalization for robust
  permutation-equivariant learning.
\newblock In {\em CVPR}, pages 11286--11295, 2020.

\bibitem{inloc}
Hajime Taira, Masatoshi Okutomi, Torsten Sattler, Mircea Cimpoi, Marc
  Pollefeys, Josef Sivic, Tomas Pajdla, and Akihiko Torii.
\newblock Inloc: Indoor visual localization with dense matching and view
  synthesis.
\newblock In {\em CVPR}, pages 7199--7209, 2018.

\bibitem{yfcc}
Bart Thomee, David~A Shamma, Gerald Friedland, Benjamin Elizalde, Karl Ni,
  Douglas Poland, Damian Borth, and Li-Jia Li.
\newblock Yfcc100m: The new data in multimedia research.
\newblock {\em Communications of the ACM}, 59(2):64--73, 2016.

\bibitem{benchmark}
Carl Toft, Will Maddern, Akihiko Torii, Lars Hammarstrand, Erik Stenborg,
  Daniel Safari, Masatoshi Okutomi, Marc Pollefeys, Josef Sivic, Tomas Pajdla,
  et~al.
\newblock Long-term visual localization revisited.
\newblock {\em TPAMI}, 2020.

\bibitem{transformer}
Ashish Vaswani, Noam Shazeer, Niki Parmar, Jakob Uszkoreit, Llion Jones,
  Aidan~N Gomez, {\L}ukasz Kaiser, and Illia Polosukhin.
\newblock Attention is all you need.
\newblock In {\em NeurIPS}, pages 5998--6008, 2017.

\bibitem{graph2}
Runzhong Wang, Junchi Yan, and Xiaokang Yang.
\newblock Learning combinatorial embedding networks for deep graph matching.
\newblock In {\em CVPR}, pages 3056--3065, 2019.

\bibitem{lowranklinformer}
Sinong Wang, Belinda~Z Li, Madian Khabsa, Han Fang, and Hao Ma.
\newblock Linformer: Self-attention with linear complexity.
\newblock {\em CoRR}, abs/2006.04768, 2020.

\bibitem{sfm2}
Changchang Wu.
\newblock Towards linear-time incremental structure from motion.
\newblock In {\em 3DV}, pages 127--134. IEEE, 2013.

\bibitem{auc1}
Kwang~Moo Yi, Eduard Trulls, Yuki Ono, Vincent Lepetit, Mathieu Salzmann, and
  Pascal Fua.
\newblock Learning to find good correspondences.
\newblock In {\em CVPR}, pages 2666--2674, 2018.

\bibitem{auc2}
Jiahui Zhang, Dawei Sun, Zixin Luo, Anbang Yao, Lei Zhou, Tianwei Shen, Yurong
  Chen, Long Quan, and Hongen Liao.
\newblock Learning two-view correspondences and geometry using order-aware
  network.
\newblock In {\em CVPR}, pages 5845--5854, 2019.

\bibitem{aachen2}
Zichao Zhang, Torsten Sattler, and Davide Scaramuzza.
\newblock Reference pose generation for visual localization via learned
  features and view synthesis.
\newblock {\em IJCV}, 129(4):821--844, 2021.

\bibitem{patch2pix}
Qunjie Zhou, Torsten Sattler, and Laura Leal-Taixe.
\newblock Patch2pix: Epipolar-guided pixel-level correspondences.
\newblock In {\em CVPR}, pages 4669--4678, 2021.

\bibitem{camerapose}
Qunjie Zhou, Torsten Sattler, Marc Pollefeys, and Laura Leal-Taixe.
\newblock To learn or not to learn: Visual localization from essential
  matrices.
\newblock In {\em ICRA}, pages 3319--3326. IEEE, 2020.

\end{thebibliography}
}
\end{document}